\documentclass{article}

\usepackage{microtype}
\usepackage{graphicx}
\usepackage{subcaption}
\usepackage{booktabs} 

\usepackage{hyperref}

\usepackage[preprint]{icml2026}

\usepackage{amsmath}
\usepackage{amssymb}
\usepackage{mathtools}
\usepackage{amsthm}

\usepackage[capitalize,noabbrev]{cleveref}

\theoremstyle{plain}
\newtheorem{theorem}{Theorem}[section]

\theoremstyle{definition}

\theoremstyle{remark}

\usepackage[textsize=tiny]{todonotes}

\icmltitlerunning{Are AI Capabilities Increasing Exponentially? A Competing Hypothesis}

\begin{document}

\twocolumn[
\icmltitle{Are AI Capabilities Increasing Exponentially? A Competing Hypothesis}



  \icmlsetsymbol{equal}{*}

  \begin{icmlauthorlist}
    \icmlauthor{Haosen Ge}{top}
    \icmlauthor{Hamsa Bastani}{sch}
    \icmlauthor{Osbert Bastani}{comp}
  \end{icmlauthorlist}

  \icmlaffiliation{top}{Wharton AI \& Analytics Initiative, The Wharton School, University of Pennsylvania, USA}
  \icmlaffiliation{sch}{Department of Operations, Information and Decisions, The Wharton School, University of Pennsylvania, USA}
  \icmlaffiliation{comp}{Department of Computer and Information Science, University of Pennsylvania, USA}

  \icmlcorrespondingauthor{Haosen Ge}{hge@wharton.upenn.edu}
  \icmlcorrespondingauthor{Hamsa Bastani}{hamsab@wharton.upenn.edu}
  \icmlcorrespondingauthor{Osbert Bastani}{obastani@seas.upenn.edu}

  \icmlkeywords{Machine Learning, ICML}

  \vskip 0.3in
]



\printAffiliationsAndNotice{}  

\begin{abstract}
Rapidly increasing AI capabilities have substantial real-world consequences, ranging from AI safety concerns to labor market consequences. The Model Evaluation \& Threat Research (METR) report argues that AI capabilities have exhibited exponential growth since 2019. In this note, we argue that the data does not support exponential growth, even in shorter-term horizons. Whereas the METR study claims that fitting sigmoid/logistic curves results in inflection points far in the future, we fit a sigmoid curve to their current data and find that the inflection point has already passed. In addition, we propose a more complex model that decomposes AI capabilities into base and reasoning capabilities, exhibiting individual rates of improvement. We prove that this model supports our hypothesis that AI capabilities will exhibit an inflection point in the near future. Our goal is not to establish a rigorous forecast of our own, but to highlight the fragility of existing forecasts of exponential growth.\footnote{Our code is available at \url{https://github.com/obastani/AI_Forecasting}.} 
\end{abstract}

\section{Introduction}

A recent report by Model Evaluation \& Threat Research (METR)~\citep{kwa2025measuring} conducted a series of analyses to measure AI capabilities in realistic tasks that require significant effort from human experts. They propose a novel metric: 50\% model horizon, which measures the difficulty of tasks that a model can solve successfully 50\% of the time. Then, they show that according to this metric, AI capabilities are increasing exponentially over time---specifically, model horizons have been doubling every seven months since 2019. Based on these results, they predict that ``within 5 years, AI systems will be capable of automating many software tasks that currently take humans a month.''

Since dissemination, this report has drawn significant attention and started heated discussion on the potential impact of such rapid improvement of AI capabilities. Much of the academic conversation has focused on safety~\citep{barnett2025ai}. However, there are substantial consequences for rapidly increasing AI capabilities beyond safety. Most notably, these results have raised substantial concerns about labor market consequences, raising the potential for large-scale displacement of skilled workers~\citep{brynjolfsson2025canaries}. Importantly, many consequences of these forecasts are immediate, shaping both policy outcomes as well as individual decisions such as choices about education and career paths. Given the substantial consequences of the potential for exponential increase in AI capabilities, there is an urgent need for rigorous methodologies for performing and validating these kinds forecasts.

\begin{figure*}
\centering
\begin{subfigure}[b]{0.45\linewidth}
\centering
\includegraphics[width=\linewidth]{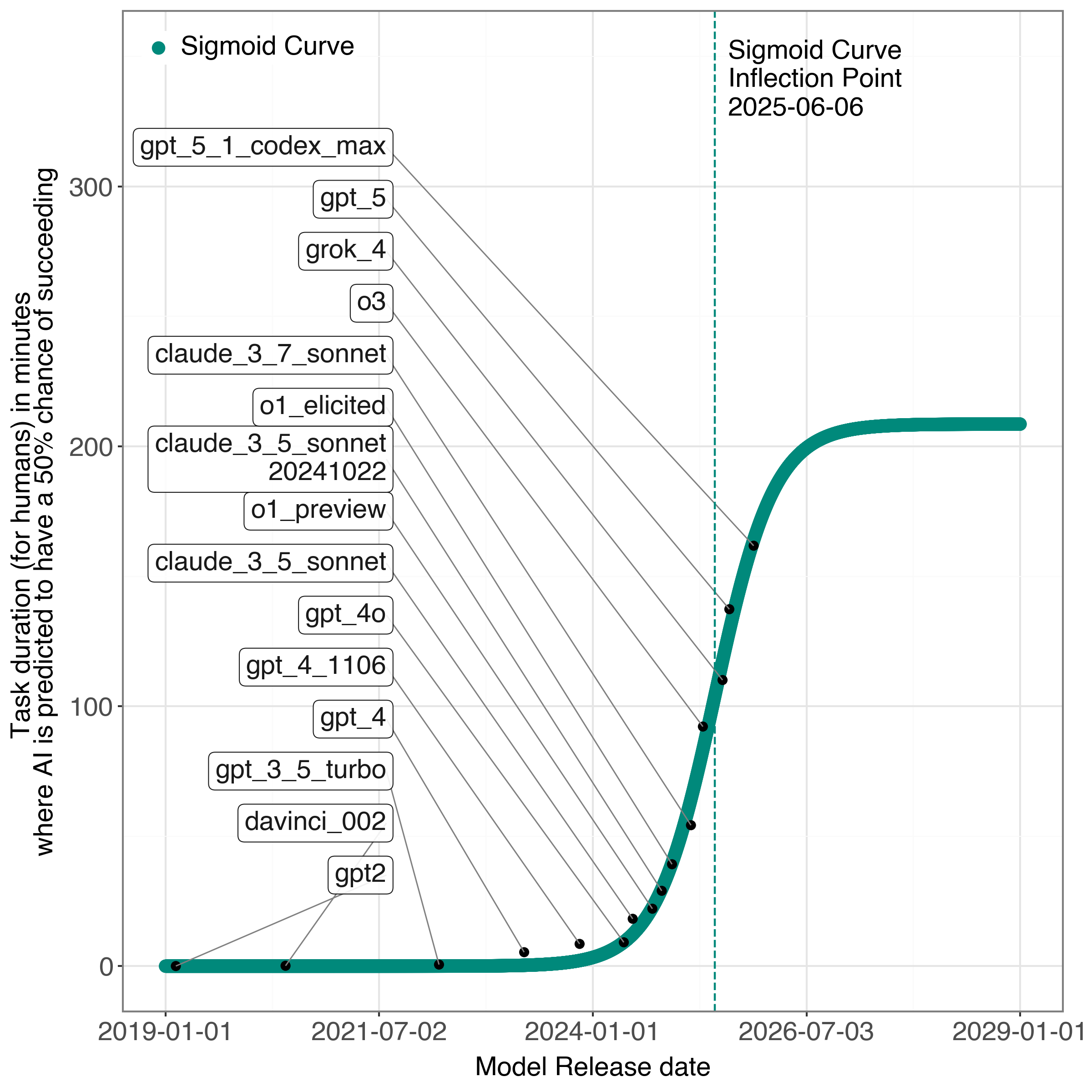}
\caption{Sigmoid Curve Fit}
\label{subfig:single_sigmoid}
\end{subfigure}%
\begin{subfigure}[b]{0.45\linewidth}
\centering
\includegraphics[width=\linewidth]{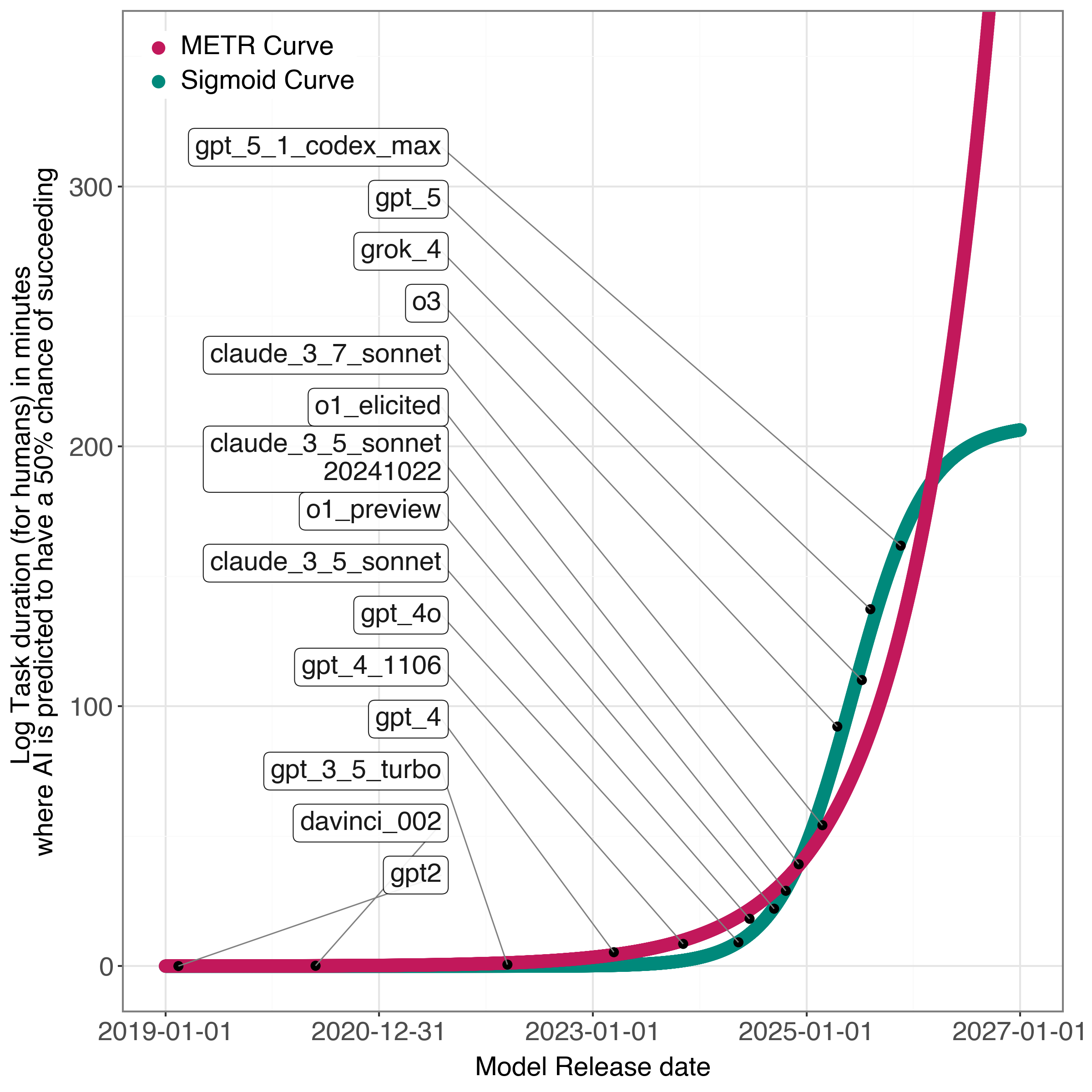}
\caption{Sigmoid Curve Fit vs METR}
\label{subfig:single_sigmoid_metr_comparison}
\end{subfigure}

\caption{\textbf{Sigmoid Curve.} This sigmoid curve is fit by minimizing the mean-squared error (MSE) of the curve $h_{\text{model}}=\gamma\cdot\sigma(\delta_1\cdot d_{\text{model}}+\delta_2)$ to the METR dataset, where $h_{\text{model}}$ is METR's ``50\% model horizon time''  for the given model, $d_{\text{model}}$ is the model release date, and $\gamma,\delta_1,\delta_2$ are parameters. We use gradient descent in PyTorch for parameter estimation. \textbf{While it is not clear that progress will plateau, recent progress clearly fits in the linear part of the sigmoid and the inflection point (2025-06-06) is in the past.}}
\label{fig:intro}
\end{figure*}

We argue to the contrary---plateauing growth is similarly well supported by the data. While the METR study compares to a small number of alternative hypotheses, aside from linear and super-exponential, none of these alternatives are visualized or discussed in detail. Most notably, they discuss the sigmoid curve as a potential alternative, but claim that their estimate of this model yields an inflection point far in the future (see Appendix~D.1 in their paper~\cite{kwa2025measuring}); they rule out plateauing growth in the near future based on this finding. However, we find that fitting a sigmoid curve results in an inflection point that is actually in the past (specifically, 2025-06-06), as visualized in Figure~\ref{fig:intro}.\footnote{While this fit is in-sample (due to the small dataset size), our main goal is to estimate the inflection point from the data rather than to provide accurate forecasts into the future.}
This finding suggests that at the very least, it is plausible that AI capabilities may plateau soon.

In general, it is impossible to rule out either alternative from data alone. Thus, to support our hypothesis, we posit a theoretical model under which the exponential appearance of recent gains in AI capabilities can be interpreted as a consequence of the introduction of reasoning capabilities into base LLMs. Specifically, we model reasoning as a separate technology that contributes multiplicatively to the overall capability of LLMs---i.e., LLM capabilities can be decomposed into two sigmoids, one for the base LLM and one for reasoning. Indeed, as can be seen from Figure~\ref{fig:intro}, the period of progress from o1-preview (released 2024-09-12) to the present forms the linear part of the sigmoid curve. Thus, our hypothesis is that following initial exponential growth due to scaling data and model size, base capabilities plateaued, but overall capabilities continued to grow for a period due to rapid improvements in reasoning.

While we present a specific alternative analysis, our goal is not to discount the METR study; in fact, we believe continuing exponential improvement is a plausible viewpoint, and it is important to take this potential outcome into consideration. However, we believe that our approach provides a plausible alternative that similarly merits consideration.

\section{Background on the METR Study}

We focus on the recent METR study~\citep{kwa2025measuring}, which forecasts that AI capabilities are exponentially increasing. This study was itself critiquing prior work, pointing out that existing metrics such as accuracy are bounded and cannot assess whether growth is exponential. To remedy this issue, they introduce a novel metric, the \emph{50\% model horizon time}, which quantifies the difficulty of tasks that a model can solve reliably. Unlike prior metrics, this one can increase unboundedly, making it suitable for assessing the possibility of exponential growth. Their analysis concludes that AI capabilities are improving exponentially.

Their experiments include three task families: HCAST, RE-Bench, and SWAA. Specifically, HCAST contains a diverse set of challenges in cybersecurity, machine learning, software engineering, and general reasoning. RE-Bench consists of challenging open-ended machine learning research engineering environments, each of which are intended to take a human expert approximately 8 hours to complete. SWAA comprises 66 small tasks commonly performed in software engineering work. In total, their study includes 170 unique tasks from the three task families. Then, they evaluate 28 popular models on the 170 tasks. Among the models, they label a subset of 15 models as state-of-the-art (SOTA), representing the frontier of AI capabilities. We reuse their experimental results and focus exclusively on the 15 state-of-the-art models to better characterize the scaling behavior of frontier AI capabilities. We include the list of SOTA models and their release dates in Table~\ref{tab:sota_model}.

\begin{table}
\centering
\caption{Release Dates of the Selected SOTA Models.}
\label{tab:sota_model}
\begin{tabular}{lc}
\toprule
\multicolumn{1}{c}{Model} & Release Date \\
\midrule
Davinci-002 & 2020-05-28 \\
GPT-4 & 2023-03-14 \\
Grok-4 & 2025-07-09 \\
Claude 3.5 Sonnet (Oct 2024) & 2024-10-22 \\
GPT-5 & 2025-08-07 \\
Claude 3.7 Sonnet & 2025-02-24 \\
GPT-2 & 2019-02-14 \\
GPT-3.5 Turbo Instruct & 2022-03-15 \\
GPT-o1-preview & 2024-09-12 \\
GPT-4 (1106) & 2023-11-06 \\
GPT-5.1 Codex Max & 2025-11-19 \\
GPT-4o & 2024-05-13 \\
Claude 3.5 Sonnet & 2024-06-20 \\
GPT-o3 & 2025-04-16 \\
GPT-o1-elicited & 2024-12-05 \\
\bottomrule
\end{tabular}
\end{table}

Their data analysis first estimates the horizon time of each model on each dataset using the following regression:
\begin{align}
p_{\text{model}} = \sigma((\log h_{\text{model}} - \log t_{\text{task}}) \cdot \beta_{\text{model}}),
\label{eq:metr_horizon}
\end{align}
where $p_{\text{model}}$ denotes the probability that the model solves a task correctly, $t_{\text{task}}$ denotes the difficulty of the task (measured by the amount of time human expert takes to complete the task), $h_{\text{model}}$ is the 50\% horizon time, $\beta_{\text{model}}$ is the parameter they estimate, and $\sigma(x)=e^x/(1+e^x)$ is the sigmoid function. By construction, $p_{\text{model}}=1/2$ when $\log h_{\text{model}} = \log t_{\text{task}}$; thus, a model with capability $h_{\text{model}}$ attains a success probability of $1/2$ on tasks with difficulty $t_{\text{task}} = h_{\text{model}}$. Thus, $h_{\text{model}}$ characterizes the task difficulty threshold at which the model achieves a success rate of $1/2$.

After estimating the 50\% horizon time, they examine the temporal trend in model capabilities by fitting the following linear regression model:
\begin{align*}
\log h_{\text{model}} = \beta_0 + \beta_1 \cdot d_{\text{model}},
\end{align*}
where $d_{\text{model}}$ denotes the model’s release date; they fit this data by treating the estimate $h_{\text{model}}$ from the previous step as ground truth $h_{\text{model}}^*$ and applying linear regression. Note that this model is equivalent to
\begin{align}
h_{\text{model}} = \exp\left(\beta_0 + \beta_1 \cdot d_{\text{model}}\right),
\label{eq:metr_final}
\end{align}
implying an exponential relationship between the model's $50\%$ horizon time and its release date. METR reports an $R^2$ value of $0.98$ for this regression, which they interpret as strong evidence that the capabilities of frontier models grow exponentially over time. In their main paper, they compare this regression to two others---linear and hyperbolic---and find that an exponential curve fits the data substantially better. They discuss additional comparisons informally in Appendix~D.1 of their paper, but do not provide quantitative evidence to support ruling out these alternatives. However, this limited comparison to alternatives makes it difficult to assess the confidence we should have in their findings.

\section{Multiplicative Model of AI Progress}
\label{sec:model}

We present a model of growing AI capabilities that decomposes it into two component technologies: the base model and reasoning. We prove that this model produces a growth curve that is qualitatively consistent with the METR data.

\subsection{Motivation}

The main hypothesis behind our model of AI progress is that much of the recent growth in AI capabilities has been driven by the introduction of reasoning into base LLMs. Specifically, while chain-of-thought reasoning has been popular for some time now~\cite{wei2022chain}, explicitly training models to perform reasoning is a more recent phenomenon, starting with OpenAI's o1 model~\cite{jaech2024openai,shao2024deepseekmath}. Since the release of o1, there has been startling progress on a number of benchmarks, with current LLMs appearing to approach the performance of human experts.

We propose a model that explicitly separates progress on base model capabilities from progress on reasoning capabilities. Intuitively, the reason this model might show support for slowing improvements is that technologies tend to exhibit very rapid growth during their introduction. Since LLMs have only been finetuned for chain-of-thought reasoning for a little over a year (starting with the introduction of OpenAI's o1 model~\cite{jaech2024openai}), it is natural that reasoning capabilities have dramatically improved capabilities over the past year. However, we might expect reasoning capabilities to start plateauing in the near future.

Without separating out improvements in reasoning, it is unsurprising that the rate of improvement appears exponential---prior to 2023, base capabilities improved exponentially due to scaling of data and model size, but these improvements plateaued due to the prohibitive cost of scaling. Reasoning capabilities have maintained this growth since 2023. Under this view, one way to interpret the METR study is that it predicts new breakthroughs will continue to prop up exponential progress. This makes sense when extrapolating, since breakthroughs have been common in the past decade; however, there is no guarantee that it will continue to be the case. Under our model, if breakthroughs stop happening, then exponential progress will end.

\subsection{Regression Model}

For our data analysis, we consider an alternative regression model to Eq.~\ref{eq:metr_final} that explicitly separates a model's base performance and its reasoning capabilities:
\begin{align*}
h_{\text{model}} &= \gamma_1 \cdot h_{\text{base}} \cdot (1 + \gamma_2 \cdot h_{\text{reasoning}})\\
h_{\text{base}} &= b(d_{\text{model}}) \\
h_{\text{reasoning}} &= r(d_{\text{model}}) \cdot 1\{k_{\text{thinking}} = 1\},
\end{align*}
where $h_{\text{model}}$ is the  50\% model horizon, $k_{\text{thinking}}$ indicates whether the model has been post-trained with reasoning capabilities and those capabilities are activated, $\gamma_1,\gamma_2$ are parameters to be estimated, and $b,r$ are link functions described below. In other words, we treat an LLM's overall capability as the product of it's base capability $h_{\text{base}}(d)$ (i.e., without reasoning) and the quality $h_{\text{reasoning}}$ of its reasoning features. Intuitively, the base capability of a LLM captures advancement of the model's pre-training phase and non-reasoning post-training phases, including model sizes and data curation, and the reasoning ability captures post-training techniques to improve chain-of-thought thinking. Both base and reasoning capabilities are functions of the model's release date $d_{\text{model}}$. The parameters $\gamma_1$ and $\gamma_2$ quantify the contributions of base capability and reasoning capability to the overall capability.

\begin{figure}
\centering
\includegraphics[width=\linewidth]{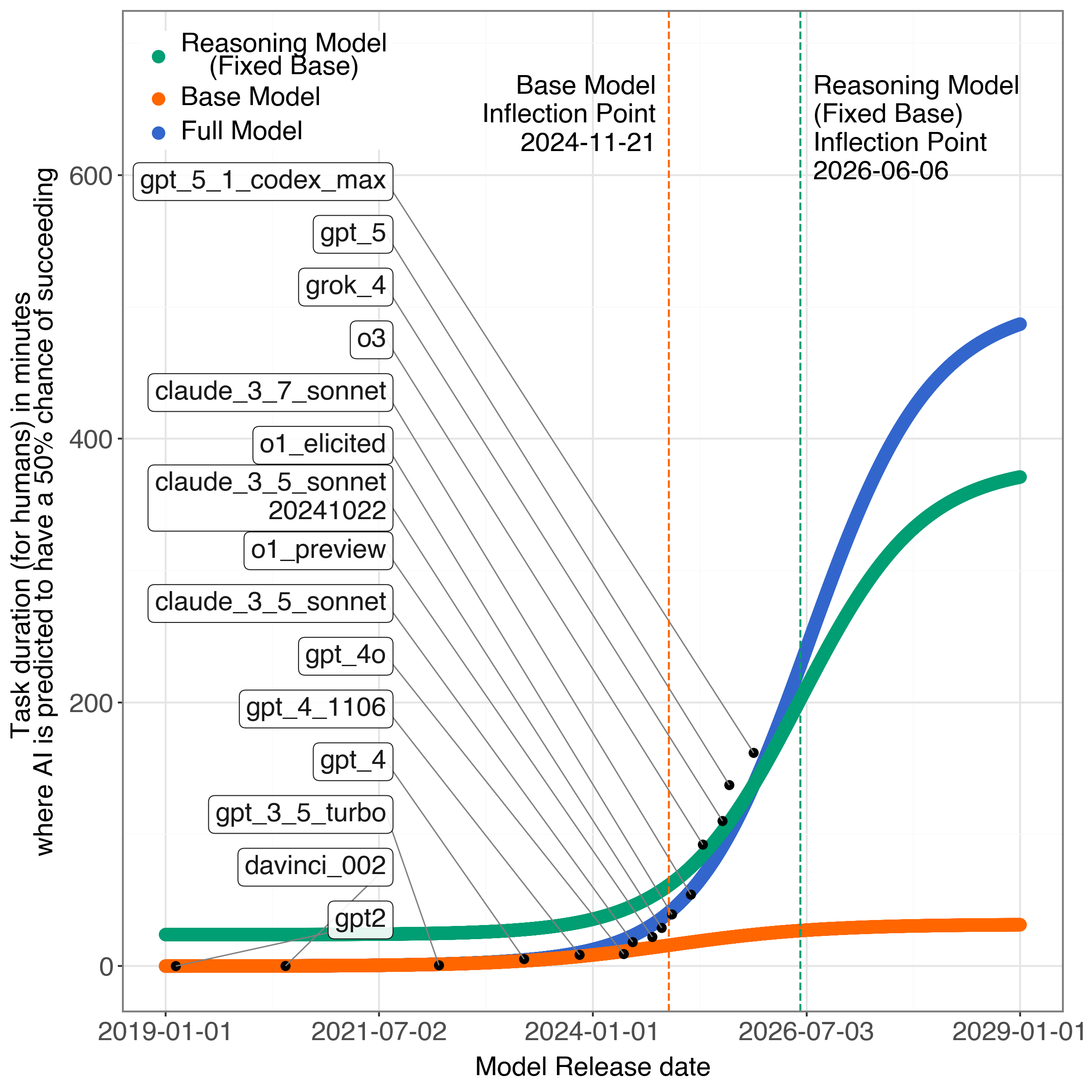}
\caption{\textbf{Sigmoid link inflection points.} The curves are as in Figure~\ref{fig:projection}; we show inflection points of the orange and green curves as dashed vertical lines of the corresponding color.}
\label{fig:sigmoid_inflection}
\end{figure}%

A key feature of our model is that technologies are \emph{multiplicative}---i.e., overall capability is the product (rather than, e.g., the sum) of the component technologies (in our case, the base model and reasoning). This assumption is the key driver behind the apparent exponential growth of staggered improvements across different technologies; we provide theoretical evidence that our model exhibits this kind of behavior in Section~\ref{sec:theory}. Intuitively, just as progress in one technology plateaus, another technology exhibits rapid improvement that props up exponential improvements. We believe this multiplicative model is realistic---for LLMs, reasoning cannot exist without strong base models, and while strong base models have useful capabilities, these are substantially boosted by reasoning.

\begin{figure*}
\centering
\begin{subfigure}[b]{0.33\linewidth}
\centering
\includegraphics[width=\linewidth]{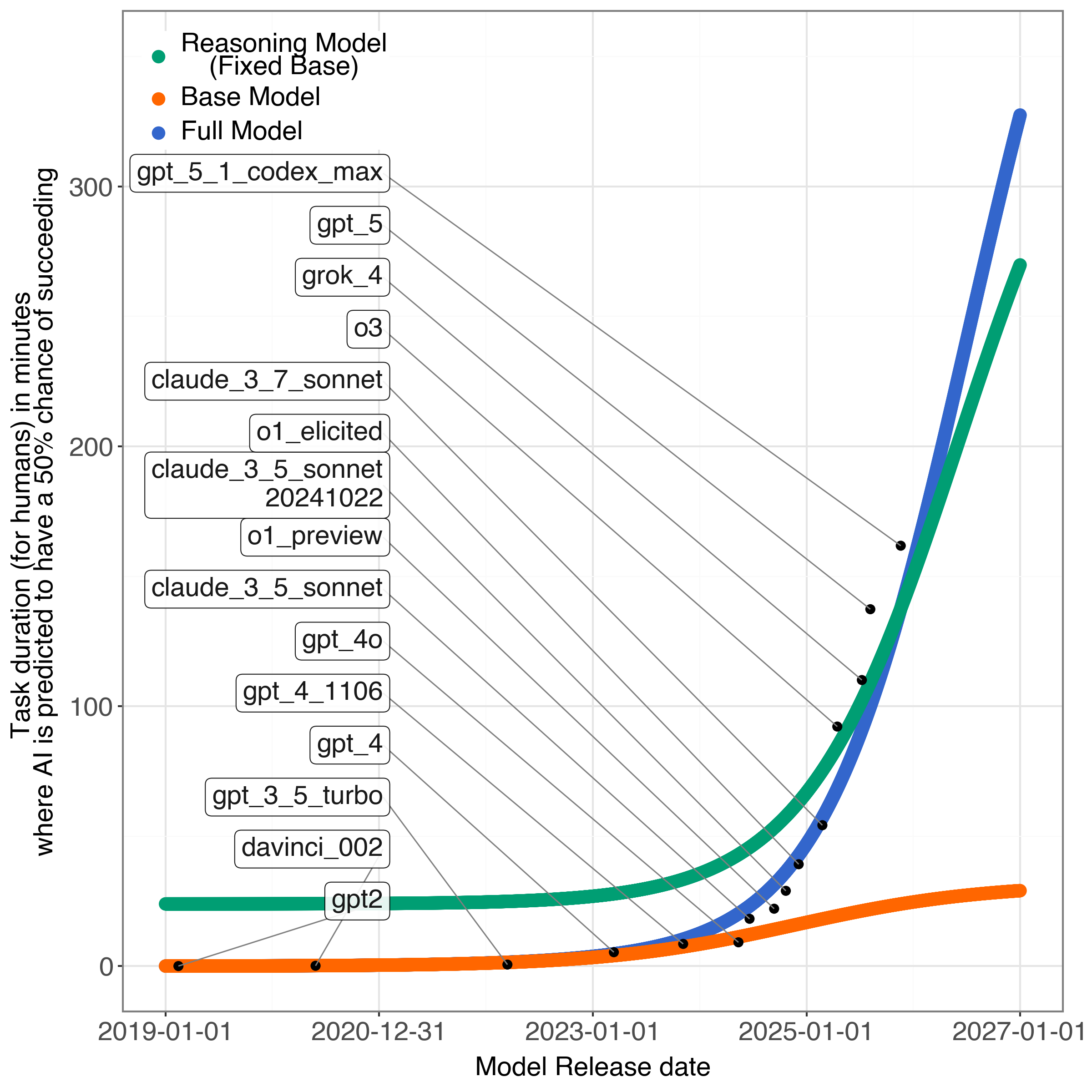}
\caption{Sigmoid Link}
\label{subfig:sigmoid}
\end{subfigure}
\hfill
\begin{subfigure}[b]{0.33\linewidth}
\centering
\includegraphics[width=\linewidth]{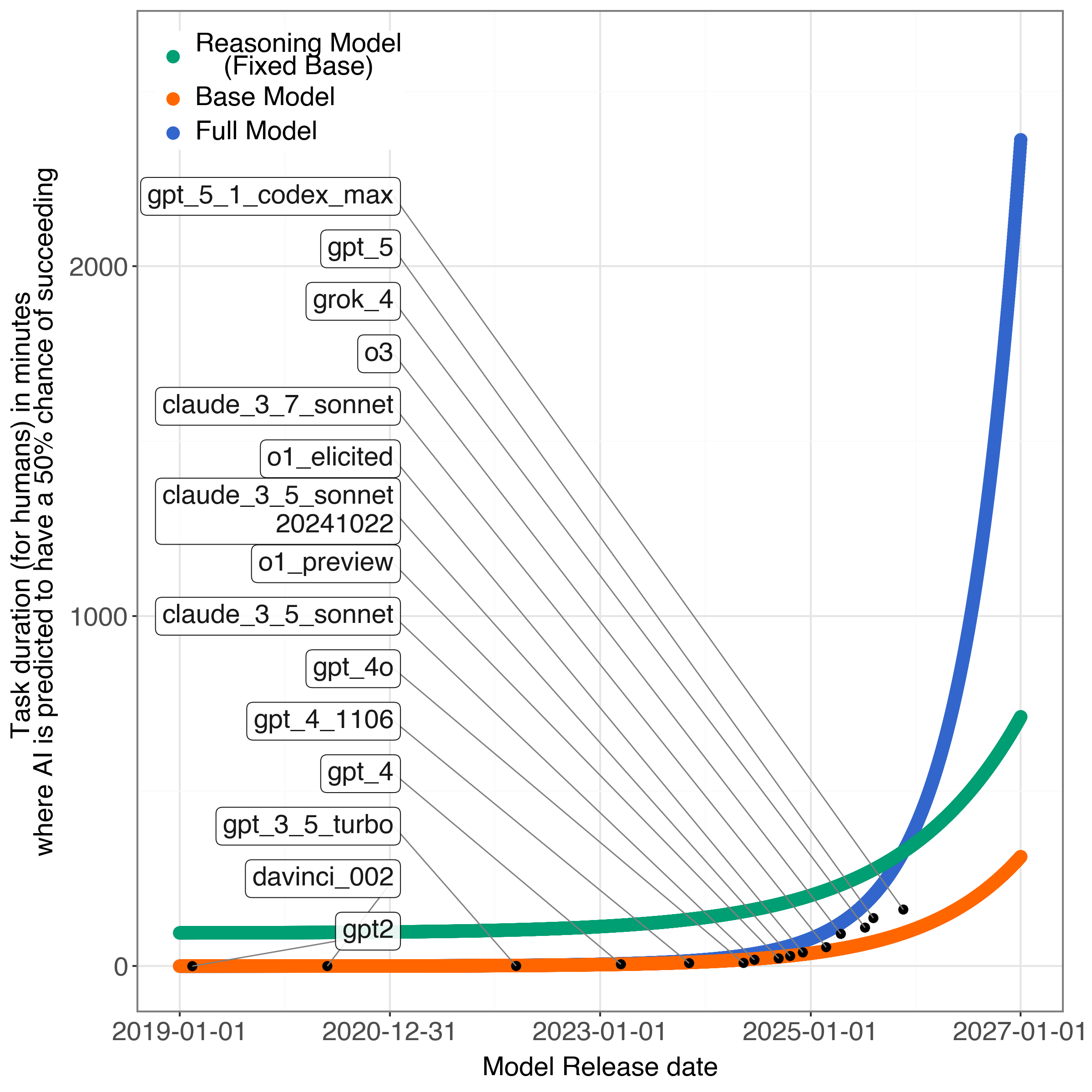}
\caption{Exponential Link}
\label{subfig:exponential}
\end{subfigure}
\begin{subfigure}[b]{0.33\linewidth}
\centering
\includegraphics[width=\linewidth]{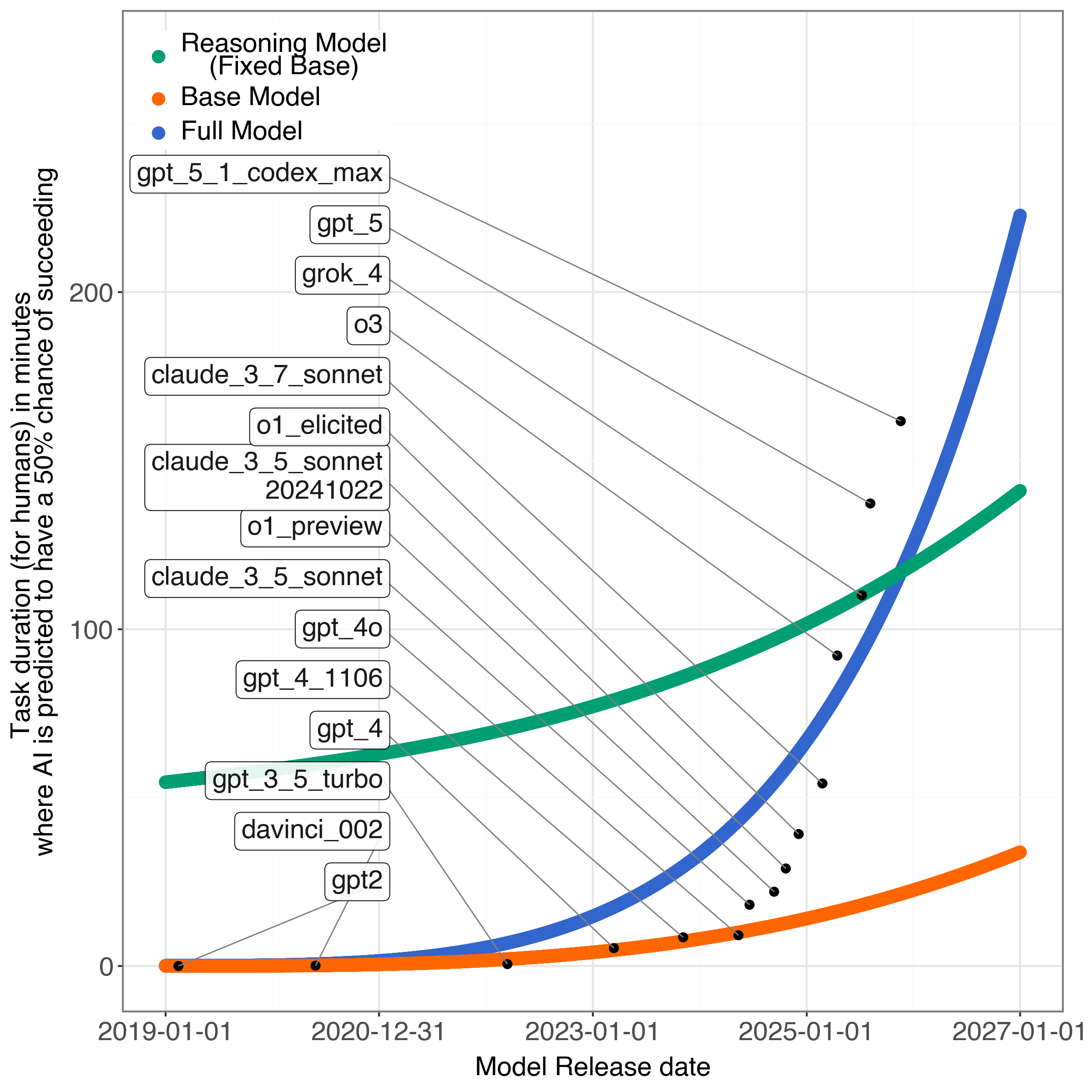}
\caption{B-Spline Link}
\label{subfig:spline}
\end{subfigure}
\hfill
\caption{\textbf{Projections under Different Link Functions.} The orange curves project base model capabilities, the green curve projects reasoning capabilities assuming the best base model (i.e., gpt-5.1-codex-max), and the blue curve shows the overall capabilities. The black points denote the 50\% model horizon estimated by METR.}
\label{fig:projection}
\end{figure*}

\subsection{Choice of Link Functions}

The link functions $b,r$ encode how the base and reasoning capabilities depend on the model's release date $d_{\text{model}}$. A wide range of functional forms could be considered; in this note, we adopt the following plausible candidates.

\textbf{Sigmoid.} First, we consider the sigmoid function:
\begin{align}
b(d_{\text{model}}) &= \sigma(\delta_1 \cdot d_{\text{model}} + \delta_2) \label{eq:f_sigmoid} \\
r(d_{\text{model}}) &= \sigma(\theta_1 \cdot d_{\text{model}} + \theta_2), \label{eq:g_sigmoid}
\end{align}
where $\delta_1,\delta_2,\theta_1,\theta_2$ are parameters. Intuitively, these link functions say that each of base and reasoning capabilities grow exponentially until reaching an ``inflection point'', after which they plateau (formally, an inflection point of an arbitrary function $f(x)$ is the point at which $f''(x)$ changes sign; the inflection of $\sigma$ is at $x=0$). In Section~\ref{sec:theory}, we provide an analysis of the implications of this model.

\textbf{Exponential.} Next, we consider an exponential function:
\begin{align*}
b(d_{\text{model}}) &= \exp(\delta_1 \cdot d_{\text{model}} + \delta_2) \\
r(d_{\text{model}}) &= \exp(\theta_1 \cdot d_{\text{model}} + \theta_2),
\end{align*}
where $\delta_1,\delta_2,\theta_1,\theta_2$ are parameters. Similar to METR's original model, these link functions say that each of base and reasoning capabilities are increasing exponentially in time; thus, the overall capability is also increasing over time.

\textbf{Spline.}
Lastly, we consider a B-spline link function:
\begin{align*}
b(d_{\text{model}}) &= \sum_{i=1}^{N_b} \delta_i\cdot B_i(d_{\text{model}}) \\
r(d_{\text{model}}) &= \sum_{i=1}^{N_r} \theta_i\cdot B_i(d_{\text{model}}),
\end{align*}
where $N_b$ and $N_r$ denote the numbers of spline basis functions for $b$ and $r$, respectively, $\{\delta_i\}_{i=1}^{N_b},\{\theta_i\}_{i=1}^{N_r}$ are parameters, and $B_i(d_{\text{model}})$ is the $i$th B-spline basis function (which we take to be a degree $m$ polynomial), which includes its own parameters. This choice of link function induces a flexible, piecewise polynomial relationship between LLM capabilities and the release date.

\subsection{Theoretical Analysis}
\label{sec:theory}

We prove that a multiplicative model of technological progress with sigmoid link exhibits exponential growth followed by plateauing. To simplify our analysis, we drop many of the parameters and assume that the core model is a product of sigmoid functions with different inflection points; we further simplify by assuming these inflection points are evenly spaced. Then, we prove that the resulting function exhibits (i) exponential growth $e^x$ before the first inflection point, (ii) squared exponential growth $e^{x^2}$ between the first and last inflection points, and (iii) plateauing thereafter.
\begin{theorem}
\label{thm:main}
Let $x$ denote time, and consider the model
\begin{align*}
f(x)=\prod_{i=1}^kf_i(x)
\quad\text{where}\quad f_i(x)=\sigma(x-i\alpha),
\end{align*}
where $\sigma$ is the sigmoid function and $\alpha\ge2$. Then:
\begin{itemize}
\item If $x\le0$, then
\begin{align*}
&\frac{1}{5}e^{kx}\exp\left(-\frac{\alpha}{2}\cdot k(k+1)\right)& \\
&\le f(x)\le e^{kx}\exp\left(-\frac{\alpha}{2}\cdot k(k+1)\right).
\end{align*}
\item If $x\in[j\alpha,(j+1)\alpha]$ for $j\in\{0,1,...,k-1\}$, then
\begin{align*}
&\frac{1}{20}\exp\left(-\frac{\alpha}{2}\cdot(k-j+1)(k-j)\right) \\
&\le f(x)
\le\exp\left(-\frac{\alpha}{2}\cdot(k-j-1)(k-j)\right).
\end{align*}
\item If $x\ge k\alpha$, then
\begin{align*}
\frac{1}{4}\le f(x)\le1.
\end{align*}
\end{itemize}
\end{theorem}
In the first case $x\le0$, the model exhibits exponential growth $e^{kx}$. In the second case, $x\approx j\alpha\approx(j+1)\alpha$, so
\begin{align*}
f(x)\approx\exp\left(-\frac{\alpha}{2}\cdot\left(k-\frac{x}{\alpha}\right)^2\right).
\end{align*}
In other words, progress continues exponentially until the final inflection point $k\alpha$. Intuitively, the exponent base decays throughout this phase; as progress for individual components plateau, the base of the exponent becomes smaller. Finally, in the third case, progress plateaus once the last inflection point has been crossed.

These trends are consistent with the data shown in Figure~\ref{fig:intro}. At a high level, the initial period of growth was exponential due to scaling AI capabilities. Many researchers believed that capabilities were plateauing; however, the introduction of post-training for reasoning sparked a second wave of improvements in capabilities, resulting in a period of steep linear increases between 2024-09-12 to the present. If our model accurately reflects reality, then new breakthroughs are necessary to sustain exponential growth.

\section{Our Analysis of the METR Data}
\label{sec:main_analysis}

We fit our model from Section~\ref{sec:model} to data; our results support the plausibility of plateauing AI capabilities.

\begin{table}
\centering
\caption{Assessment of goodness of fit via MSE on $h_{\text{model}}$.}
\label{tab:mse}
\begin{tabular}{cr} 
\toprule Specification & \multicolumn{1}{c}{Mean Squared Error (MSE)} \\ 
\midrule 
Sigmoid Link & 203.69 \\ 
B-Spline Link & 511.80 \\ 
Exponential Link & 2874.67 \\ 
\midrule
METR Exponential Curve & 339.93 \\ 
Sigmoid Curve & 27.37 \\
\bottomrule
\end{tabular}
\end{table}

\begin{figure*}
\centering
\begin{subfigure}[b]{0.45\linewidth}
\centering
\includegraphics[width=\linewidth]{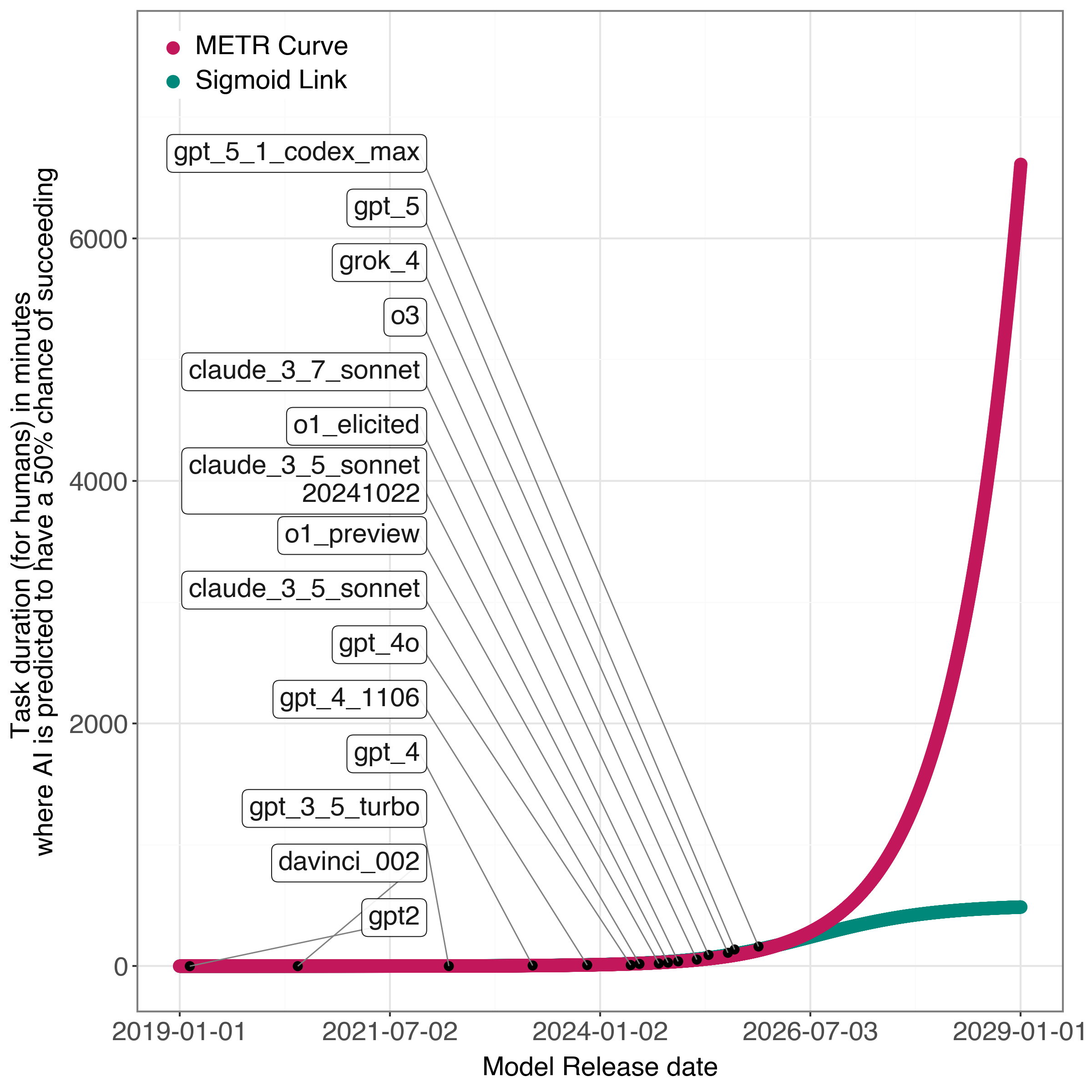}
\caption{Long-term Projection}
\label{subfig:comparison_long_term}
\end{subfigure}%
\begin{subfigure}[b]{0.45\linewidth}
\centering
\includegraphics[width=\linewidth]{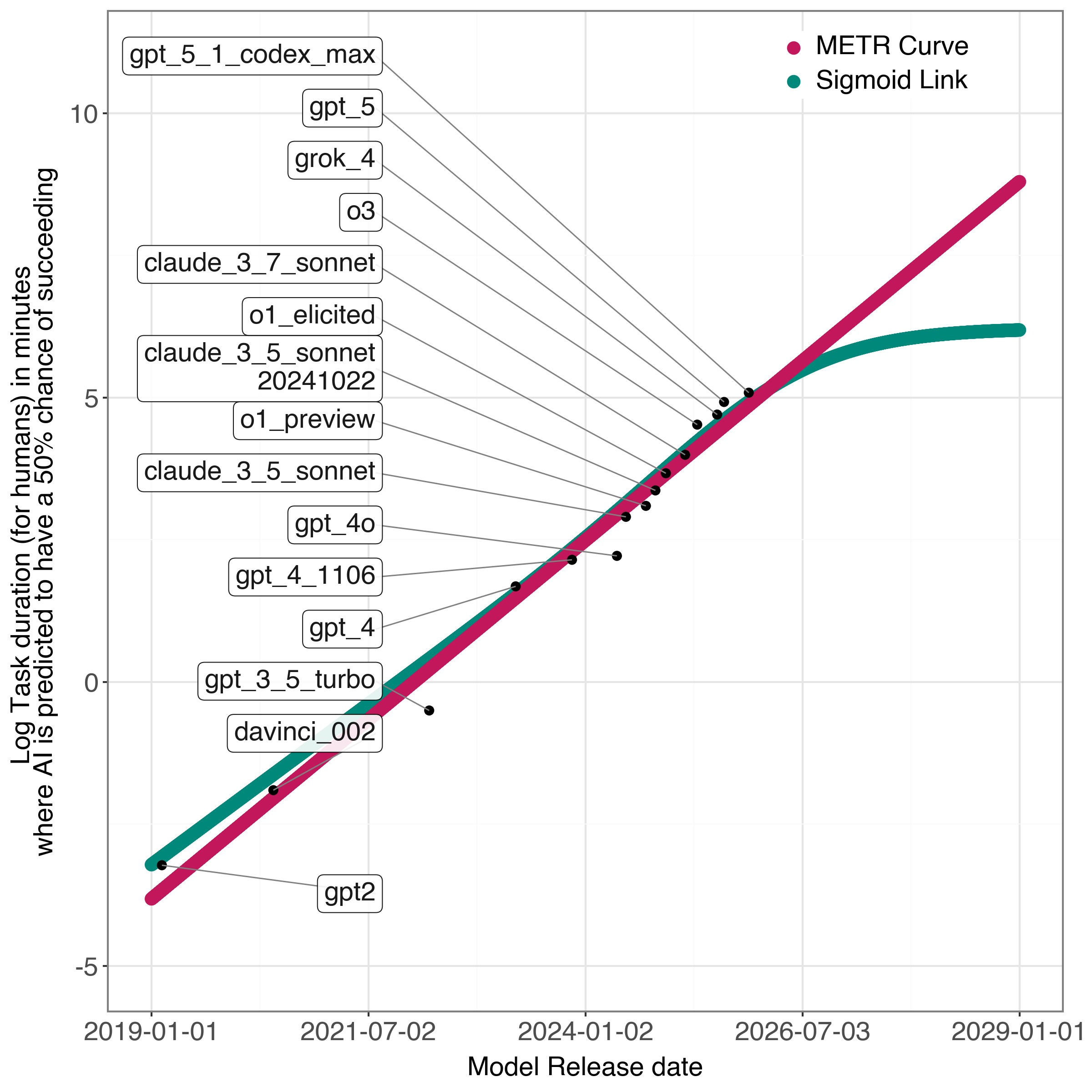}
\caption{Log Model Capability}
\label{subfig:log_model_capability}
\end{subfigure}
\caption{\textbf{Comparison of Sigmoid Link and METR Projection.}}
\label{fig:sigmoid_metr_comparison}
\end{figure*}

\subsection{Methodology}

We estimate our models the original experiment data shared by the METR study~\citep{kwa2025measuring} (including data from the HCAST, RE-Bench, and SWAA benchmarks). We access the data via METR's public Github repository: \url{https://github.com/METR/eval-analysis-public}.

We perform estimation by maximizing the log-likelihood of the probabilistic model $p_{\text{model}}$ in Eq.~\ref{eq:metr_horizon} using \texttt{Stan}~\citep{carpenter2017stan}. For the sigmoid and exponential link functions, we adopt the weakly informative prior $\mathcal{N}(0, 10^2)$ for all parameters, and impose positivity constraints on $\gamma_1, \gamma_2, \delta_1, \theta_1,\beta_{\text{model}}$. For the B-spline link function, we use two breakpoints and polynomial splines of degree five for both $b$ and $r$ (i.e., degree $m = 5$ and $N_b = N_r = 2 + 5 - 1 = 6$). To ensure that $h_{\text{model}}$ remains strictly positive and to avoid taking the logarithm of a negative quantity, we constrain all spline coefficients $\delta_i$ and $\theta_i$ to be positive. Furthermore, we regularize the spline coefficients using random-walk priors, following standard practice to mitigate overfitting: 
$\delta_1 \sim \mathcal{N}(0,1)$, $\delta_i \sim \mathcal{N}(\delta_{i-1}, \tau)$ for all $i \geq 2$, and $\tau \sim \mathcal{N}(0,1)$. The same priors are applied to $\theta_i$.

\subsection{Results}

\textbf{Model Fit.} Figure~\ref{fig:projection} shows the fitted models;  that combine base and reasoning components. Given the limited number of available models, we assess goodness-of-fit using the in-sample mean squared error (MSE) between the model-predicted AI capability and the observed AI capability: $(h_{\text{model}}-h_{\text{model}}^*)^2$; Table~\ref{tab:mse} reports the resulting MSEs. We also include METR's exponential curve and our sigmoid curve in Figure~\ref{fig:intro}; note that these models are not directly comparable since they optimize different loss functions. Among the three link functions considered, the sigmoid-link model achieves by far the lowest MSE, suggesting that within our model, sigmoid growth appears more plausible than exponential growth. Our approach also outperforms METR's exponential curve, though our model has more parameters so this comparison is not rigorous.


\textbf{Inflection points.}
The key question is not the magnitude of recent performance gains (which are undeniable), but whether these gains will continue. One argument in the METR study is that there is no sign of an ``inflection point'' where exponential increase starts to slow down; we have already shown in Figure~\ref{fig:intro} that the current data does not support this claim. To further understand inflection points in AI capabilities, we plot the inflection points of Eqs.~\eqref{eq:f_sigmoid} \&~\eqref{eq:g_sigmoid} in Figure~\ref{fig:sigmoid_inflection}. The inflection points $d_b$ of $b(d)$ and $d_r$ of $r(d)$ based on the estimated parameters are
\begin{align*}
\hat{d}_b = \text{2024-11-21}
\qquad\text{and}\qquad
\hat{d}_r = \text{2026-06-06},
\end{align*}
respectively. These results support our finding---the inflection point $\hat{d}_b$ for base model capabilities happened just after the release of the first reasoning model (o1-preview) in 2024-09-12, and the inflection point $\hat{d}_r$ for reasoning capabilities is projected to happen in the near future. Thus, increasing reasoning capabilities drove the substantial overall improvements we have seen in the past two years.

Baked into our model is the idea that unless significant breakthroughs happen, progress will plateau. Thus, the question about forecasting AI capabilities becomes a question of whether we should expect another breakthrough that produces improvements on the same scale as reasoning. This question can only be answered by domain knowledge; we leave it for the broader community to answer.

These results highlight an additional benefit of decomposing overall capability into component technologies---it enables us to understand and interpret forecasts of progress for individual components separately, providing an understanding of \emph{why} progress looks a certain way.

\textbf{Long-term forecasts.}
Finally, we compare METR’s exponential forecast with our sigmoid-link model over a ten-year horizon from 2019-01-01 to 2029-01-01. Results are shown in Figure~\ref{subfig:comparison_long_term}. The two models yield similar projections up to approximately 2026-07-03. Beyond this point, the METR model predicts an increasingly rapid rise in model capability, whereas our model suggests that capabilities will plateau in the near future. External forecasts cannot in general be validated, so these results highlight the importance of leveraging domain insights to assess the validity of forecasts.


\section{Limitations}
\label{sec:limitations}

\textbf{In-sample evaluation.} A key limitation is that our estimates are all evaluated in-sample; while the same is true for existing methodologies such as the METR study, our models also include more parameters. This limitation is inevitable due to the limited amount of data available. However, our goal is not to provide irrefutable evidence that AI capabilities are plateauing, but that it is a highly plausible alternative to continuing exponential growth. As more data becomes available, it is critical to assess which models have more accurately forecast progress to build confidence in future forecasts. More broadly, more rigorous methodologies must be developed for assessing the accuracy of these forecasts.

\textbf{Evaluation metric.} A related issue is that comparing across different kinds of models (ours vs. METR) is complicated by the fact that they are estimated in very different ways. The METR paper performs regression to minimize the MSE in the space of log-outcomes $\log h_{\text{model}}$, whereas we have used probabilistic modeling directly on the final outcomes $p_{\text{model}}$; furthermore, our sigmoid curve in Figure~\ref{fig:intro} minimizes MSE on outcomes $h_{\text{model}}$. While we have used MSE at predicting $h_{\text{model}}$ to compare models, this comparison is not fair due to the diverse loss functions, especially given the in-sample comparison. Again, this issue necessitates the development of more rigorous evaluation methodologies.

\textbf{Multiplicative assumption.} The multiplicative assumption is critical for driving our theoretical analysis. While we believe this assumption presents at least a plausible alternative, substantial work needs to be done to validate it in practice.

\textbf{Limited decomposition.} We have only modeled base and reasoning capabilities; we believe these are the key driving factors behind the substantial recent gains in AI capabilities, but there are many other important factors. Prior to the release of GPT-4, many of the gains in performance were driven simply by scaling the amount of training data and the model size, as well as basic instruction tuning. Recent improvements in reasoning capabilities have been driven by improved post-training procedures and the creation of datasets tailored to reasoning tasks; other non-public improvements may also have contributed. In principle, these two components can be further decomposed into aspects such as data engineering, pre- and post-training algorithms, network architecture, etc.; however, we currently lack enough information about state-of-the-art LLMs to perform a more granular analysis. Finding ways to address these issues could help lead to more accurate forecasts.

\section{Conclusion}

To contrast with the recent view on exponential increase of LLM capabilities, we have presented an alternative viewpoint arguing that this increase is not exponential; instead, it is plateauing or possibly linear. Our forecasting methodology combines domain-specific modeling that decomposes progress into separate base and reasoning capabilities, with an empirical analysis based on the METR dataset, which we believe could form the basis of future forecasting models. Importantly, our methodology is primarily intended to be a compelling alternative rather than a definitive rebuttal. We believe that substantially more work needs to be done both in terms of improved forecasting methodologies, as well as improved evaluation of these methodologies.

\bibliography{ref}

\begin{thebibliography}{14}
\providecommand{\natexlab}[1]{#1}
\providecommand{\url}[1]{\texttt{#1}}
\expandafter\ifx\csname urlstyle\endcsname\relax
  \providecommand{\doi}[1]{doi: #1}\else
  \providecommand{\doi}{doi: \begingroup \urlstyle{rm}\Url}\fi

\bibitem[Barnett \& Scher(2025)Barnett and Scher]{barnett2025ai}
Barnett, P. and Scher, A.
\newblock Ai governance to avoid extinction: The strategic landscape and actionable research questions.
\newblock \emph{arXiv preprint arXiv:2505.04592}, 2025.

\bibitem[Brynjolfsson et~al.(2025)Brynjolfsson, Chandar, and Chen]{brynjolfsson2025canaries}
Brynjolfsson, E., Chandar, B., and Chen, R.
\newblock Canaries in the coal mine? six facts about the recent employment effects of artificial intelligence.
\newblock \emph{Digital Economy}, 2025.

\bibitem[Carpenter et~al.(2017)Carpenter, Gelman, Hoffman, Lee, Goodrich, Betancourt, Brubaker, Guo, Li, and Riddell]{carpenter2017stan}
Carpenter, B., Gelman, A., Hoffman, M.~D., Lee, D., Goodrich, B., Betancourt, M., Brubaker, M., Guo, J., Li, P., and Riddell, A.
\newblock Stan: A probabilistic programming language.
\newblock \emph{Journal of statistical software}, 76:\penalty0 1--32, 2017.

\bibitem[Ho et~al.(2025)Ho, Denain, Atanasov, Albanie, and Shah]{ho2025rosetta}
Ho, A., Denain, J.-S., Atanasov, D., Albanie, S., and Shah, R.
\newblock A rosetta stone for ai benchmarks.
\newblock \emph{arXiv preprint arXiv:2512.00193}, 2025.

\bibitem[Jaech et~al.(2024)Jaech, Kalai, Lerer, Richardson, El-Kishky, Low, Helyar, Madry, Beutel, Carney, et~al.]{jaech2024openai}
Jaech, A., Kalai, A., Lerer, A., Richardson, A., El-Kishky, A., Low, A., Helyar, A., Madry, A., Beutel, A., Carney, A., et~al.
\newblock Openai o1 system card.
\newblock \emph{arXiv preprint arXiv:2412.16720}, 2024.

\bibitem[Jimenez et~al.(2023)Jimenez, Yang, Wettig, Yao, Pei, Press, and Narasimhan]{jimenez2023swe}
Jimenez, C.~E., Yang, J., Wettig, A., Yao, S., Pei, K., Press, O., and Narasimhan, K.
\newblock Swe-bench: Can language models resolve real-world github issues?
\newblock \emph{arXiv preprint arXiv:2310.06770}, 2023.

\bibitem[Kwa et~al.(2025)Kwa, West, Becker, Deng, Garcia, Hasin, Jawhar, Kinniment, Rush, Von~Arx, et~al.]{kwa2025measuring}
Kwa, T., West, B., Becker, J., Deng, A., Garcia, K., Hasin, M., Jawhar, S., Kinniment, M., Rush, N., Von~Arx, S., et~al.
\newblock Measuring ai ability to complete long tasks.
\newblock \emph{arXiv preprint arXiv:2503.14499}, 2025.

\bibitem[Maslej et~al.(2025)Maslej, Fattorini, Perrault, Gil, Parli, Kariuki, Capstick, Reuel, Brynjolfsson, Etchemendy, et~al.]{maslej2025artificial}
Maslej, N., Fattorini, L., Perrault, R., Gil, Y., Parli, V., Kariuki, N., Capstick, E., Reuel, A., Brynjolfsson, E., Etchemendy, J., et~al.
\newblock Artificial intelligence index report 2025.
\newblock \emph{arXiv preprint arXiv:2504.07139}, 2025.

\bibitem[Owen(2025)]{epochepri2025aipower}
Owen, D.
\newblock What will ai look like in 2030?, 2025.
\newblock URL \url{https://epoch.ai/files/AI_2030.pdf}.

\bibitem[Patwardhan et~al.(2025)Patwardhan, Dias, Proehl, Kim, Wang, Watkins, Fishman, Aljubeh, Thacker, Fauconnet, et~al.]{patwardhan2025gdpval}
Patwardhan, T., Dias, R., Proehl, E., Kim, G., Wang, M., Watkins, O., Fishman, S.~P., Aljubeh, M., Thacker, P., Fauconnet, L., et~al.
\newblock Gdpval: Evaluating ai model performance on real-world economically valuable tasks.
\newblock \emph{arXiv preprint arXiv:2510.04374}, 2025.

\bibitem[Shao et~al.(2024)Shao, Wang, Zhu, Xu, Song, Bi, Zhang, Zhang, Li, Wu, et~al.]{shao2024deepseekmath}
Shao, Z., Wang, P., Zhu, Q., Xu, R., Song, J., Bi, X., Zhang, H., Zhang, M., Li, Y., Wu, Y., et~al.
\newblock Deepseekmath: Pushing the limits of mathematical reasoning in open language models.
\newblock \emph{arXiv preprint arXiv:2402.03300}, 2024.

\bibitem[Sinha et~al.(2025)Sinha, Arun, Goel, Staab, and Geiping]{sinha2025illusion}
Sinha, A., Arun, A., Goel, S., Staab, S., and Geiping, J.
\newblock The illusion of diminishing returns: Measuring long horizon execution in llms.
\newblock \emph{arXiv preprint arXiv:2509.09677}, 2025.

\bibitem[Wei et~al.(2022)Wei, Wang, Schuurmans, Bosma, Xia, Chi, Le, Zhou, et~al.]{wei2022chain}
Wei, J., Wang, X., Schuurmans, D., Bosma, M., Xia, F., Chi, E., Le, Q.~V., Zhou, D., et~al.
\newblock Chain-of-thought prompting elicits reasoning in large language models.
\newblock \emph{Advances in neural information processing systems}, 35:\penalty0 24824--24837, 2022.

\bibitem[Wijk et~al.(2024)Wijk, Lin, Becker, Jawhar, Parikh, Broadley, Chan, Chen, Clymer, Dhyani, et~al.]{wijk2024re}
Wijk, H., Lin, T., Becker, J., Jawhar, S., Parikh, N., Broadley, T., Chan, L., Chen, M., Clymer, J., Dhyani, J., et~al.
\newblock Re-bench: Evaluating frontier ai r\&d capabilities of language model agents against human experts.
\newblock \emph{arXiv preprint arXiv:2411.15114}, 2024.

\end{thebibliography}
\bibliographystyle{icml2026}

\appendix

\section{Related Work}

\textbf{Benchmarking LLMs.}
A number of complex evaluation frameworks have been proposed to assess AI performance under realistic conditions. SWE-bench evaluates LLMs on real-world GitHub pull requests paired with gold-standard fixes and unit tests~\citep{jimenez2023swe}. GDPval compiles a comprehensive set of tasks designed to be representative of the U.S. economy~\citep{patwardhan2025gdpval}; domain experts from each representative industry are recruited to design tasks for AI systems, while additional experts evaluate model outputs and annotate whether AI-generated solutions are preferred over human completions. \citet{ho2025rosetta} propose a method for stitching together existing benchmarks to assess long-term AI performance. \citet{wijk2024re} introduce RE-bench, which collects challenging research-and-development tasks hand-crafted by human experts.

\textbf{Forecasting AI capabilities.}
\citet{maslej2025artificial} systematically documents advances and trends in AI across multiple domains over time, providing an empirical basis for forecasting. Epoch AI has published papers and reports offering in-depth analyses of AI capabilities and future development trajectories (e.g., \citet{epochepri2025aipower}). \citet{sinha2025illusion} distinguish between ``execution'' and ``planning,'' focusing on a model’s ability to correctly execute a complex but predefined plan. They show that diminishing improvements in single-step accuracy can compound, resulting in exponential growth in the length of tasks a model can complete.

\section{Proof of Theorem~\ref{thm:main}}
\label{sec:proof}

First, suppose that for some $j\in\{0,1,...,k-1\}$, we have $x\in[j\alpha,j'\alpha]$, where $j'=j+1$. For all $i\le j$,
\begin{align*}
f_i(x)
=1-\frac{e^{-x+i\alpha}}{1+e^{-x+i\alpha}}
\ge1-\frac{e^{-(j-i)\alpha}}{2},
\end{align*}
so we have
\begin{align*}
\prod_{i=1}^jf_i(x)
&\ge1-\frac{1}{2}\sum_{i=1}^je^{-(j-i)\alpha}
\ge1-\frac{1}{2}\sum_{h=0}^{\infty}e^{-h\alpha} \\
&\ge1-\frac{1}{2}\sum_{h=0}^{\infty}\frac{1}{3^h}
\ge\frac{1}{4},
\end{align*}
since $e^{-2}\le1/3$. In addition, we have $f_i(x)\le1$, so $\prod_{j=1}^if_i(x)\le1$. Next, for all $i\ge j'$, we have
\begin{align*}
f_i(x)
=\sigma(x-i\alpha)
=\frac{e^{x-i\alpha}}{1+e^{x-i\alpha}}
\ge\frac{e^{-(i-j)\alpha}}{1+e^{-(i-j')\alpha}},
\end{align*}
so
\begin{align*}
\prod_{i=j'}^kf_i(x)
\ge\frac{\prod_{i=j'}^ke^{-(i-j)\alpha}}{\prod_{i=j'}^k(1+e^{-(i-j')\alpha})}.
\end{align*}
For the numerator, we have
\begin{align*}
\prod_{i=j'}^ke^{-(i-j)\alpha}
&=\prod_{h=1}^{k-j}e^{-h\alpha}
=\exp\left(-\alpha\sum_{h=1}^{k-j}h\right) \\
&=\exp\left(-\frac{\alpha}{2}\cdot(k-j+1)(k-j)\right).
\end{align*}
For the denominator, we have
\begin{align*}
&\prod_{i=j'}^k(1+e^{-(i-j')\alpha})
\le\prod_{h=0}^{\infty}(1+e^{-h\alpha}) \\
&=\exp\left(\sum_{h=0}^{\infty}\log(1+e^{-h\alpha})\right)
\le\exp\left(\sum_{h=0}^{\infty}e^{-h\alpha}\right) \\
&\le\exp\left(\frac{1}{1-e^{-\alpha}}\right)
\le5.
\end{align*}
In addition, $f_i(x)\le e^{x-i\alpha}\le e^{-(i-j')\alpha}$, so
\begin{align*}
\prod_{i=j'}^kf_i(x)
&\le\exp\left(-\frac{\alpha}{2}\cdot(k-j')(k-j)\right).
\end{align*}
Putting everything together, we have
\begin{align*}
&\frac{1}{20}\exp\left(-\frac{\alpha}{2}\cdot(k-j+1)(k-j)\right) \\
&\le f(x)
\le\exp\left(-\frac{\alpha}{2}\cdot(k-j-1)(k-j)\right).
\end{align*}
Next, if $x\ge k\alpha$, a similar argument shows that
\begin{align*}
\frac{1}{4}\le f(x)\le1.
\end{align*}
Finally, if $x\le0$, then we have
\begin{align*}
\prod_{i=1}^kf_i(x)
\ge\frac{\prod_{i=1}^ke^{x-i\alpha}}{\prod_{i=1}^k(1+e^{x-i\alpha})}.
\end{align*}
For the numerator, we have
\begin{align*}
\prod_{i=1}^ke^{x-i\alpha}
&=e^{kx}\exp\left(-
\alpha\sum_{i=1}^ki\right) \\
&=e^{kx}\exp\left(-\frac{\alpha}{2}\cdot k(k+1)\right),
\end{align*}
and the denominator is bounded by $5$ as before. In addition,
\begin{align*}
\prod_{i=1}^kf_i(x)
\le\prod_{i=1}^ke^{x-i\alpha}
&\le e^{kx}\exp\left(-\alpha\sum_{i=1}^ki\right) \\
&=e^{kx}\exp\left(-\frac{\alpha}{2}\cdot k(k+1)\right).
\end{align*}
Thus, we have
\begin{align*}
&\frac{1}{5}e^{kx}\exp\left(-\frac{\alpha}{2}\cdot k(k+1)\right)& \\
&\le f(x)\le e^{kx}\exp\left(-\frac{\alpha}{2}\cdot k(k+1)\right).
\end{align*}
The claim follows. $\qed$

\end{document}